# Unraveling the Impact of Initial Choices and In-Loop Interventions on Learning Dynamics in Autonomous Scanning Probe Microscopy

Boris N. Slautin[1,a)], Yongtao Liu[2], Hiroshi Funakubo[3], Sergei V. Kalinin[4,a)]

[1] Independent researcher, Belgrade, 11000, Serbia

[2] Center for Nanophase Materials Sciences, Oak Ridge National Laboratory, Oak Ridge, TN 37831, USA

[3] Department of Material Science and Engineering, Tokyo Institute of Technology, Yokohama 226-8502, Japan.

[4] Department of Materials Science and Engineering, University of Tennessee, Knoxville, TN 37996, USA

**ABSTRACT**

The current focus in Autonomous Experimentation (AE) is on developing robust workflows to conduct the AE effectively. This entails the need for well-defined approaches to guide the AE process, including strategies for hyperparameter tuning and high-level human interventions within the workflow loop. This paper presents a comprehensive analysis of the influence of initial experimental conditions and in-loop interventions on the learning dynamics of Deep Kernel Learning (DKL) within the realm of AE in Scanning Probe Microscopy. We explore the concept of 'seed effect', where the initial experiment setup has a substantial impact on the subsequent learning trajectory. Additionally, we introduce an approach of the seed point interventions in AE allowing the operator to influence the exploration process. Using a dataset from Piezoresponse Force Microscopy (PFM) on $PbTiO_3$ thin films, we illustrate the impact of the 'seed effect' and in-loop seed interventions on the effectiveness of DKL in predicting material properties. The study highlights the importance of initial choices and adaptive interventions in optimizing learning rates and enhancing the efficiency of automated material characterization. This work offers valuable insights into designing more robust and effective AE workflows in microscopy with potential applications across various characterization techniques.

---

[a)] Authors to whom correspondence should be addressed: bslautin@gmail.com and sergei2@utk.edu

# I. INTRODUCTION

In the last few years, autonomous experiment (AE) has become progressively more important in materials science [1–11]. Various methodologies have been designed to automate the synthesis of materials spanning from organic molecules and inorganic nanoparticles to metal halide perovskites [7–9,12–14]. In turn, substantial efforts are directed toward developing AE for local characterization techniques, including electron microscopy [15,16], scanning probe microscopy (SPM) [17–26], neutron diffraction [27,28], and X-ray scattering [29].

Historically, AE was hindered by the need for equipment modifications to engender remote or automated control. Reflecting the disruptive changes in the community, today the design of self-driven systems is supported by the gradual integration of the Python application programming interfaces (APIs) into operational equipment [2]. Similarly, the development of user-friendly programming libraries wrapping the powerful machine learning (ML) algorithms increases the affordability of advanced ML tools [30,31]. Additionally, user-friendly Python libraries improve the attractivity of AE, enabling its application in solving routine research tasks. The current significant challenge lies in developing robust workflows – the sequence of operations, data analysis, decision-making, etc. – to conduct the AE effectively [2].

The noticeable progress in automatization has been achieved in scanning probe microscopy [11]. Here, the ML agents proved their efficiency in automating tasks such as image analysis, correction, and segmentation [24,32–34]. Implementations of ML agents to monitor the SPM probe condition through *in-situ* scan analysis were recently demonstrated [22,25]. Additionally, several automated approaches have been showcased for identifying and exploring molecules and molecular clusters adsorbed on the surface [23,26], and surface point defects [21].

Recently, we demonstrated the AE workflow for the discovery of structure-property relations based on myopic optimization for microscopy applications [19]. A. Biswas et al. introduced the Bayesian Optimized Active Recommender System (BOARS), enabling the dynamic construction of the experiment target through a human operator's real-time voting for acquired data spectra[35]. The core of these workflows often involves Deep Kernel Learning (DKL) – a neural network employed to convert high-dimensional input structural data to low-dimensional descriptors, thereby enabling Gaussian Process/Bayesian optimization (GP/BO) for decision-making [36]. This DKL-based workflow has been successfully applied to Piezoresponse Force Microscopy (PFM) [17,19], conductive Atomic Force Microscopy (cAFM) [37], Scanning

Transmission Electron Microscopy Electron Energy Loss Spectroscopy (STEM-EELS) [16], and 4D STEM [15].

Further improvements in workflow also require well-defined approaches to control the AE process, including strategies for hyperparameters tuning and opportunities for high-level human interventions in the workflow loop [2,35]. Previously, we have proposed methodologies for the estimation of the learning rate of the DKL-based AE and approaches for monitoring the progression of the AE via analysis of learning curves, and real and feature spaces [38]. Here we explore the influence of the experiment initialization (*seed effect*) and seed points interventions, where the locations are determined not by the AE but via an alternative sampling model, on the learning dynamic of DKL-based AE.

The exploration of the seed effect and seed points interventions is carried out on the band excitation piezoresponse spectroscopy (BEPS) model dataset collected from a $PbTiO_3$ (PTO) thin film [19]. The dataset consists of the local hysteresis loops sampled on a uniform grid within the BE PFM scan area. The main objective of the autonomous experiment in PFM is to reveal the correlation between the pre-existed structural data and specific target functionality. This enables the reconstruction of the distribution target functionality across the PFM scan. For BEPS data, the PFM image plays the role of structural data, while investigating any local physical properties associated with the shape of hysteresis loops can be deemed as a target.

The DKL workflow for the SPM measurements can be described as an iterative four-stage process (Fig. 1a-d). At the beginning of DKL AE, we have access only to the global PFM image (structural image). In the first stage, the global image is represented as a collection of microstructural patches ($X_*$) – small square regions centered at each individual location $[x, y]$ within the field of view (Fig. 1a,f). Following this, it is required to select a few microstructural patches $(i, j)$ to conduct initial spectra measurements at the corresponding locations $[x_i, y_i]$ (*seed points*). The second stage of the DKL workflow is devoted to the formation of prior knowledge for the DKL training (Fig. 1b). Here we implement the *scalarizer function* to extract interesting characteristics from the spectrums collected in the seed points. The extracted characteristics are interrelated with the aimed physical property. In our case of BEPS data, such hysteresis loop parameters as area (work of switching), height (switchable polarization), build-in bias (imprint charge), etc. can be used as the scalarizers [39] (Fig. 1e).

We train the DKL by the available pairs of seed structural patches and specific *scalarizers*, utilizing them as inputs and targets correspondently (Fig. 1f). Note that all structural patches and only several seed scalarizers are available at the beginning of the experiment. The trained DKL is implemented to predict both the values and uncertainties of the scalarizer functionality across all structural patches. In the final step, Bayesian optimization is used to determine the next location for spectrum measurement (Fig. 1d). For this, a classical acquisition function – such as upper confidence bound (UCB), expected improvement (EI), and maximum uncertainty (MU) – is constructed based on acquired predictions and uncertainties. The next location is derived from the position of the maximum/minimum of the acquisition function. The seed dataset is expanded by incorporating the newly collected spectrum and structural patch and process iterated from stage two while the predefined goal is achieved, or the experimental budget is exhausted. A more detailed explanation of the primary stages and general principles of DKL AE along with potential post-experiment forensic analysis was recently published by us [38].

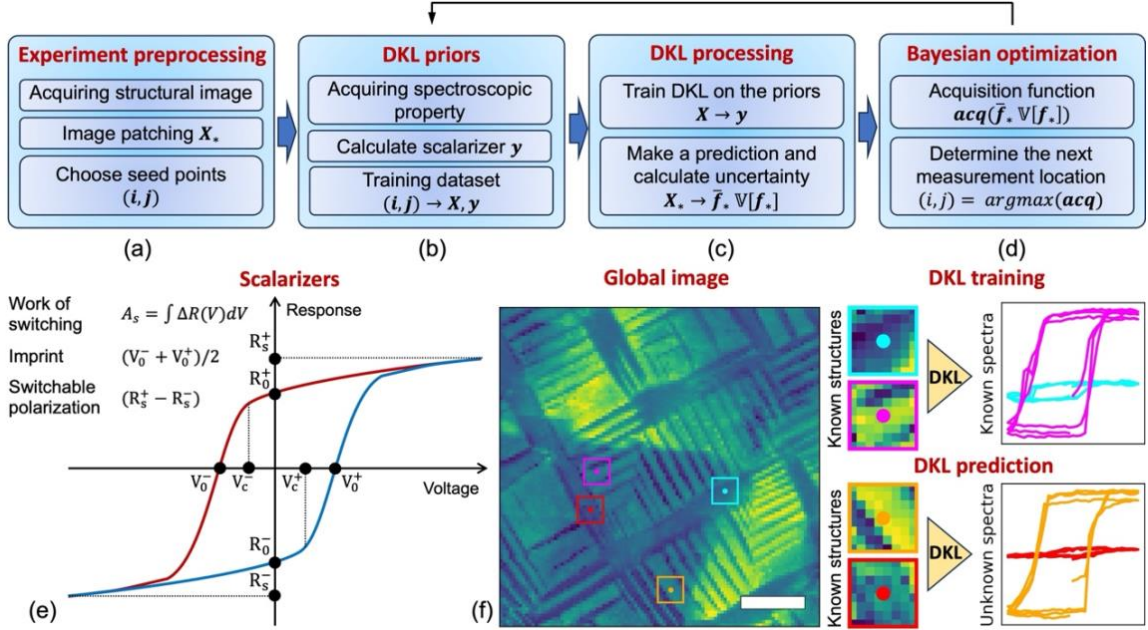

**FIG. 1.** Schematic illustration of the DKL autonomous experiment for PFM: (a-d) the main stages of the DKL workflow, (e) scalarizers available from the PFM hysteresis loop, (f) representation of the DKL training by the pairs of known microstructural patches and spectral data, and representation of the unknown spectra (scalarizer value) prediction by the trained DKL. The scale bar is 600 nm.

## II. CHALLENGES IN DKL AUTOMATED EXPERIMENT

As described above, the DKL AE is a complex multi-stage decision process driven by the acquisition function. Throughout the experiment, uncertainties obtained at each iteration can be used for the examination of the efficiency of exploration or injection of unknowns, e.g., when DKL continuously learns similar features, the uncertainty is expected to decline, while injections of new features (that are unknown to DKL) may lead to an abrupt jump of uncertainty. Another powerful tool to monitor the exploration dynamic is an *experimental trace* – the sequence of measuring locations, determined by the DKL BO, and both scalarizer and full spectra acquired at these locations [38].

Conducting dynamic analysis using the experimental trace constructed over the real space (across the global image) can potentially be challenging due to the complexity of interpreting variability factors among high-dimensional image patches. Implementing (rotationally invariant) variational autoencoders (rVAE) to the entire set of microstructural patches allows to rectify this problem by disentangling variability factors [40,41]. This approach transforms the patches into points within a two-dimensional latent space (latent distribution). The distance between these points in the VAE latent space is proportional to the dissimilarity of the corresponding microstructures. Therefore, the configuration and local densities of points inside the latent distribution encapsulate information about the structural diversity. Observing an *experimental trace* both in the real and VAE latent space allows us to comprehensively estimate the exploration dynamic.

As a preprocessing for the DKL experiment, the pre-existing BE PFM image (Fig. 2a) was divided into a series of microstructural patches, each measuring 8x8 pixels in size. We applied rVAE encoding to this set of structural patches. The obtained values of the latent variables $z_1$ and $z_2$ (Fig. 2b,c) only particularly correlate with the structural data, emphasizing various aspects of the microstructure's variability. It is important to note, that the latent variables do correspond to variations in physical features inside the scan (domains, domain walls, etc.), but this correlation can be very complex for human interpretation [42,43]. Nevertheless, we can speculate that in our dataset $z_2$ represents more of the global changes in the response level, while $z_1$ is interrelated with the local domains and domain wall contrasts. The points located in regions of higher latent distribution density correspond to the more statistically valuable microstructural patches (Fig. 2d).

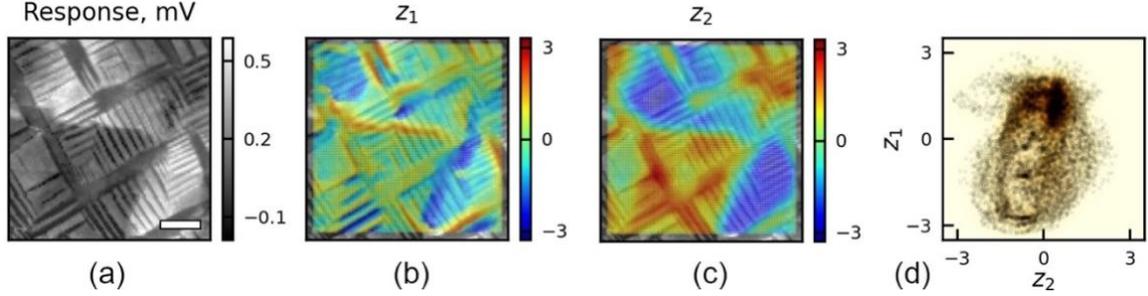

**FIG. 2.** rVAE representation of the structural data: (a) global image, (b,c) $z_1$ and $z_2$ latent variables, (c) latent distribution. The scale bar is 600 nm.

In our AE experiment simulation, the scalarizer was chosen to be the area of the hysteresis loops, related to the work of polarization switching. We initiated the AE exploration multiple times, employing EI, MU, and UCB acquisition functions. For each initiation, we randomly selected 5 seed points among the structural patches to get scalarizer functionality there. The progress of DKL exploration was estimated by the *learning curves* representing the average DKL uncertainty and its variation following each exploration step (Fig. 3a,d,h). Through forensic analysis of the learning curves after the initial 50 exploration steps (DKL iterations), we uncovered two distinct scenarios in the DKL experiment evolution: one characterized as "normal" and the other as "exploratory stagnation". It is important to emphasize that both scenarios were observed for each of the acquisition functions. The examples of the learning curves for them are shown by the violin plots in Fig. 3.

In the case of a "normal" evolution scenario, the average uncertainty exhibits substantial variability but generally trends downward with exploration (Fig. 3a,d,h red graphs). It is important to highlight that for the MU acquisition function, which solely focuses on exploration, the learning curve shows a more consistent and stable decrease throughout the process (Fig. 3d) compared to the learning curves for EI and UCB acquisition functions (Fig. 3a,h). We associated considerable redistribution of the kernels within the violin plots particularly during the initial steps with active exploration. The volume of new information gained at each step at the beginning of exploration is compatible with the pre-acquired knowledge, causing significant fluctuations in predictions and uncertainties between consecutive steps. The uncertainty fluctuations decrease with the step number, but some "sharp peaks" within the violine plot kernels are observed at the entire presented training range. These peaks are associated with the unexpected discoveries of atypical and rare microstructures during the automated experiment, leading to the appearance of enormous DKL uncertainty in these points. The analysis of the experimental traces, both in real and latent spaces, also demonstrates active exploration during the initial stages of the experiment (Fig. 3b,e,i). Measurement locations are generally distributed across the entire global image in the real space. However, a minor clustering near specific points is observed for the EI and USB acquisition functions, indicating exploitation components in them (Fig. 3b,i). In the rVAE latent space, the density of points, representing the microstructural patches chosen by DKL AE (i.e., measurement locations), correlates with the overall point density of the entire latent distribution.

Oppositely, we obtained a consistently decreasing and smooth DKL learning curve for the "exploratory stagnation" scenario, regardless of the seed point distributions and acquisition functions used (Fig. 3a,d,h blue graphs). The DKL uncertainty possesses minimal variability over the entire presented training range. The examination of real experimental traces indicates that measurement locations tend to cluster around specific points within the global image, neglecting further exploration of the real space. An analogous trapping effect was also observed within the experimental traces mapped onto the rVAE latent space (Fig. 3c,f,j). The selected by the AE measurement locations correspond to the microstructures represented by the points in the periphery of the rVAE latent distribution, while central regions, including the most statistically valuable areas, remain unexplored. Interestingly, these specific points are the same in all experiments, indicating the existence of the local maximum of all acquisition functions in them.

There are several ways of human intervention in the loop of automated experiment workflow to address the "exploratory stagnation". The human interventions can involve adjusting the scalarizer (e.g. transition from loop area to width or height), correcting the balance between exploration and exploitation, tuning the hyperparameters of the acquisition function, or even altering the acquisition function itself [38]. Additionally, the selection of initial measurement locations and the interruption of the AE exploration through seed point intervention may influence the decision-making process. These seed points initialization and interventions can rectify the evolutionary scenarios within the experiment. This highlights the potential for discovering more case-specific and sensible models for choosing the measurement location for DKL priors and interventions to enhance the learning rate and improve resilience against experiment trapping.

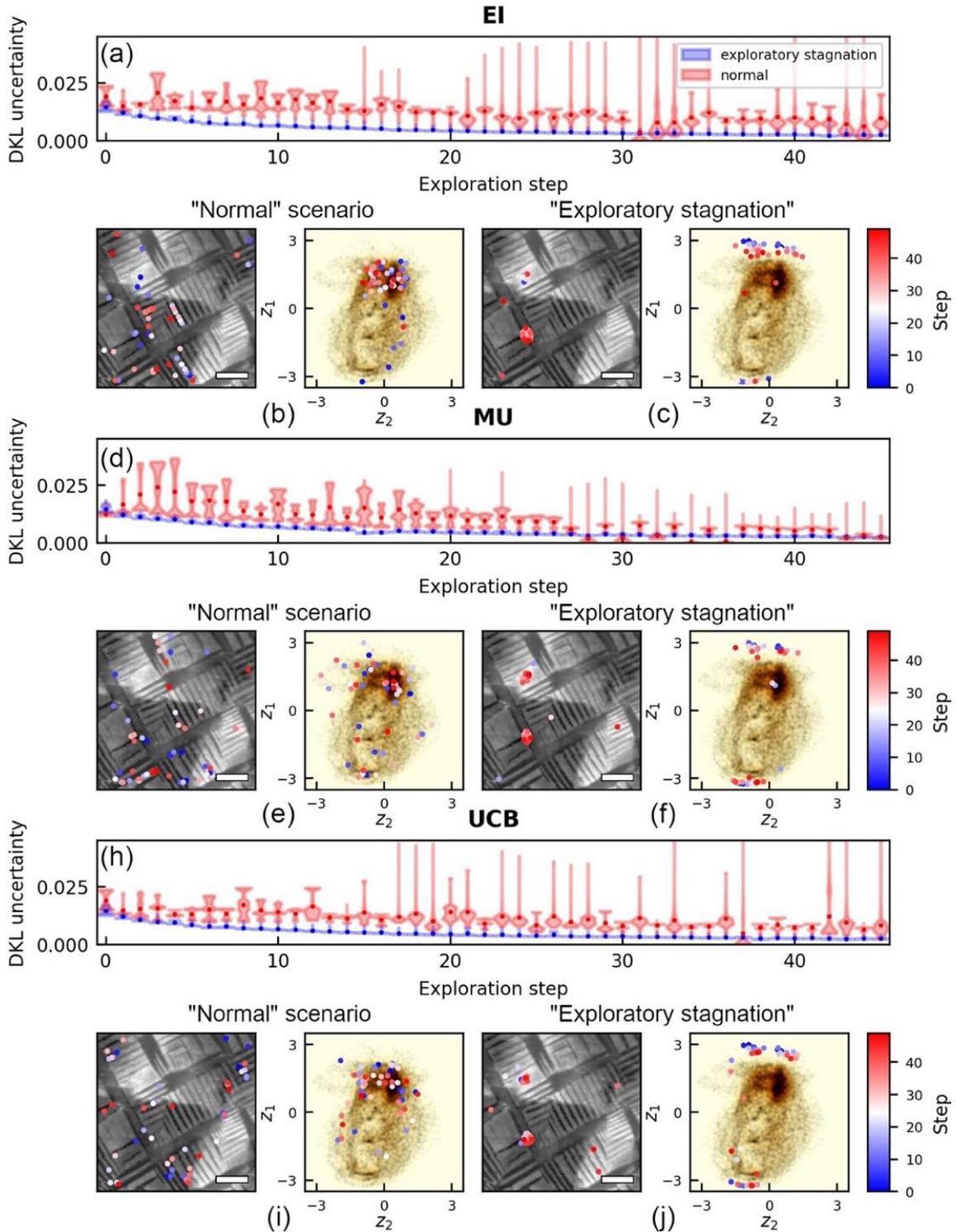

**FIG. 3.** Examples of DKL AE: (a, d, h) learning curves represented by violine plot for EI, MU, UCD acquisition functions, correspondently; experimental traces mapped onto real space and rVAE latent space (b, e, i) for "normal" scenario and (c, f, j) for "exploratory stagnation". The scale bar is 600 nm.

## III. SEED POINTS SOLUTIONS

The representation of the microstructural data within the rVAE latent space offers a comprehensive insight into their diversity and statistical significance within the global image. We utilized rVAE latent space as a foundation to construct three probabilistic models for the sampling:

1) Gaussian kernel density estimation (KDE) over the latent space distribution – **GD** (Fig. 4a).
2) Uniform probability across the rVAE latent distribution – **UD** (Fig. 4b).
3) Uniform probability across the entire rVAE latent space – **ULS** (Fig. 4c).

The Gaussian KDE is built over the latent space distribution of the real microstructures, so employing the GD model is analogous to a random selection of the seed points. The likelihood of selecting specific microstructures is proportional to the local point density in a correspondent region of the latent space (Fig. 4d). Put differently, the microstructure types that are more prevalent in the global image will be chosen more frequently compared to rare microstructures.

In the UD and ULS models selected points are spread uniformly across the latent distribution region (Fig. 4e) or the entire latent space (Fig. 4f), correspondently. This occurs regardless of the statistical significance of the corresponding microstructural patches, prioritizing the diversity of the selected microstructures.

The algorithm of points selection consists of two stages. In the first step, a chosen sampling model is used to acquire the required number of seed points. In most instances, the obtained points do not correspond to the real microstructures from the global image. To address this, we choose the nearest points from the latent distribution to the acquired ones (Fig. 4g,h,i). The part of the seed points selected through the ULS model might fall beyond the boundaries of the latent distribution (Fig. 4f), resulting in the closest points from the distribution's edge (Fig. 4i). We pose, that UD and ULS models can exhibit similar behavior for small subsamplings, but with increasing subsampling size ULS prioritize the microstructures from the distribution edges. It should be noted that the extent of edge prioritizing can vary and depends on the specific shape of the latent distribution. Additionally, as latent distribution is a subsampling derived from Gaussian KDE, the procedure of selecting the closest points leads to the GD model influencing the ULS and UD seed point selections.

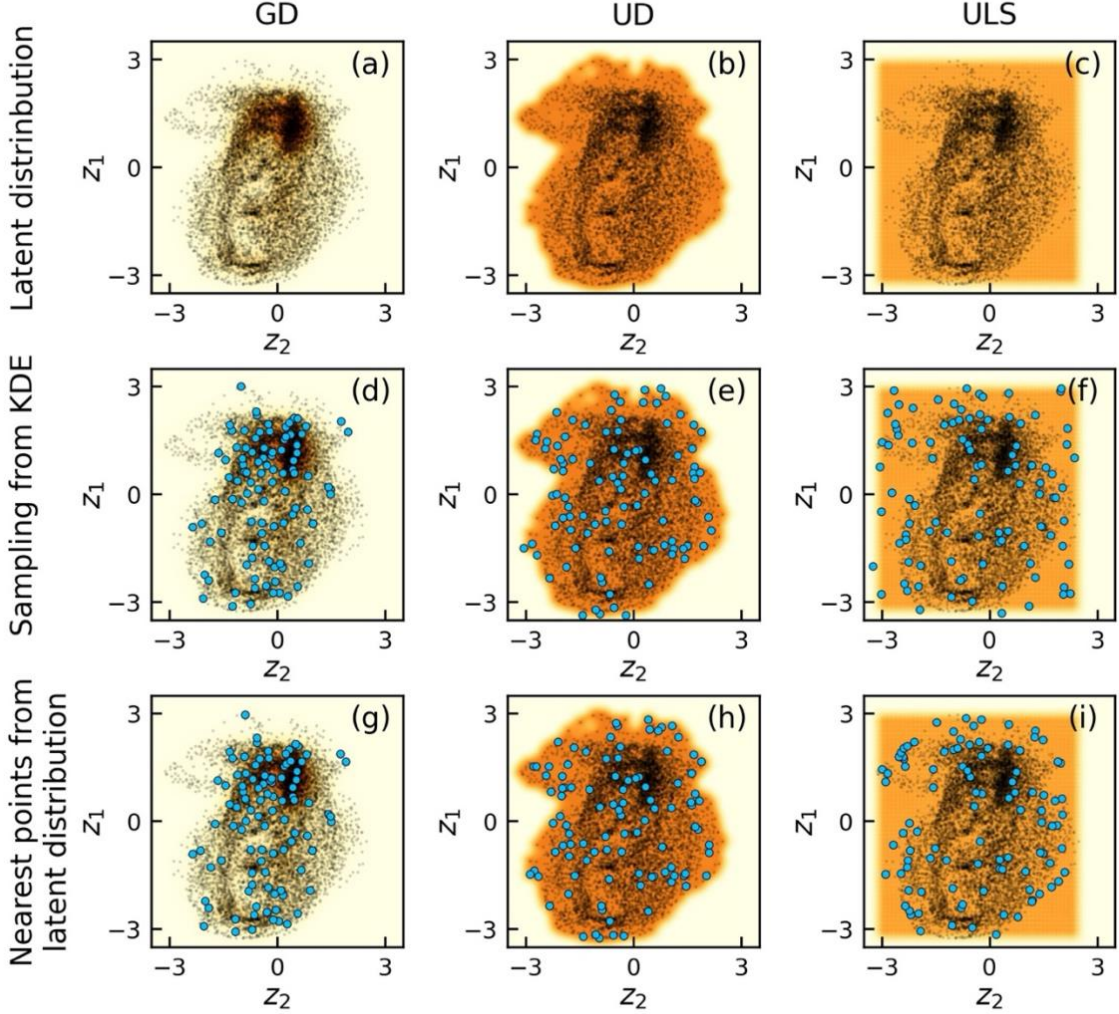

**FIG. 4.** Initial sampling models: (a-c) latent distributions with probability densities, (d-f) 100 points sampled by respective models, (g-i) nearest points from the latent distributions to the sampled ones.

We have trained DKL using various numbers of initial points, acquired through the GD, UD, and ULS models to estimate their efficiency. The trained models were used to predict the scalarizer functionality across the global image. The quality of prediction was estimated by the average DKL uncertainty and mean absolute error (MAE). We repeated this experiment 50 times for each model to obtain statistical insights and represented results in the form of violin plots (Fig. 5). The points inside the violin kernels represent the mean DKL uncertainty averaged across both the feature space and experiments.

The smooth decline observed in the average DKL uncertainty and average MAE indicates information accumulation with the increment in measurement points. Simultaneously, some deviations in violin kernel sizes for DKL uncertainty plots were observed, especially for the ULS and UD sampling models. Above ten seed points, the GD model exhibits relatively smaller kernel sizes compared to the other models. This highlights an enhanced variability of resulting seed point distributions across the feature space for ULS and UD sampling models. We speculate that the extent of difference among the distinct sampling models relies on the variety of microstructures within the exploring space. So, differences between the GD, UD, and ULS models might be significantly more pronounced in real experiments involving a larger number of measurement locations and greater microstructural diversity.

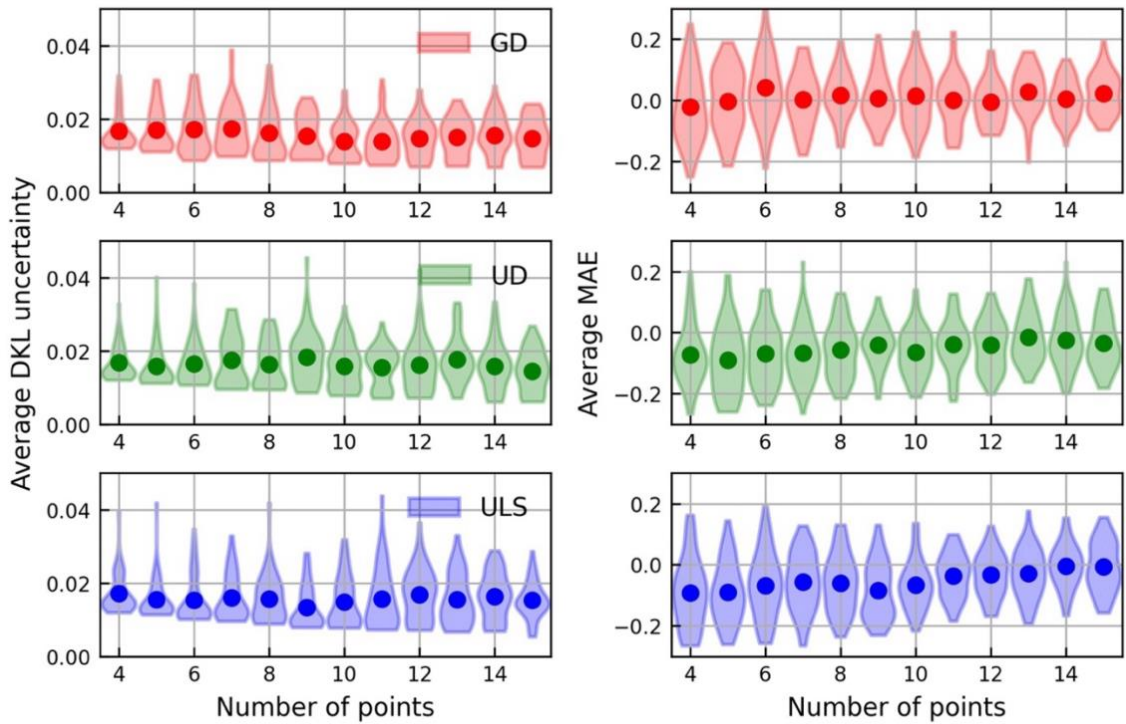

**FIG. 5.** Average DKL uncertainty and MAE vs number of seed points initialized with GD, UD, and ULS sampling models.

## IV. SEED POINTS INITIALIZATION EFFECT

To discover the influence of the different seed point initializing sampling policies on AE, we conducted 45 experiment simulations – 15 for each of the GD, UD, and ULS models. All experiments are started from the 5 seed points selected by the specific model. The resulting learning curves were categorized based on their association with either the "normal" or "exploratory stagnation" scenarios to analyze separately (Fig. S1). The DKL uncertainties were averaged across both the feature space and "normal" experiments and represented in Fig. 6a-c. The variabilities of average DKL uncertainty across experiments are illustrated in Fig. 6d-f. Additionally, we carried out 15 DKL experiment simulations wherein each subsequent measurement location was chosen randomly (Fig. 6, represented by black line).

We observe the differences between average DKL uncertainty for "normal" AE started with various sampling models during the first ~10-20 steps, suggesting that the seed points choice holds significance within this initial interval. The variation of the average DKL uncertainty typically demonstrates some growth at the first exploration steps, where the influence of seed points is crucial (Fig. 6d-f). The average DKL uncertainty shows lower values for experiments started with UD model for the MU acquisition function (Fig. 6b). The ULS sampling model statistically exhibits higher values of DKL uncertainty compared with the other ones (Fig. 6a-c). It can be related to the tendency to choose the most rare and untypical microstructures from the latent distribution edges. The GD sampling policy shows lower DKL uncertainty when employing the EI acquisition function (Fig. 6a). The learning curves showcase similar behaviors as the curve resulting from random point selection for all acquisition functions. The DKL uncertainty curves estimated for the AE with subsequent point selection conducted randomly decrease at the same rate as for AE guided by BO at the initial exploration stage. It is crucial to note that the average DKL uncertainty at the initial stage of the experiment cannot directly estimate AE efficiency. However, it does reveal distinctions in the experiment's dynamics and aids in estimating the stages where initial sampling models affect evolution. The extent of influence from the initial point selection could significantly vary based on the specific dataset.

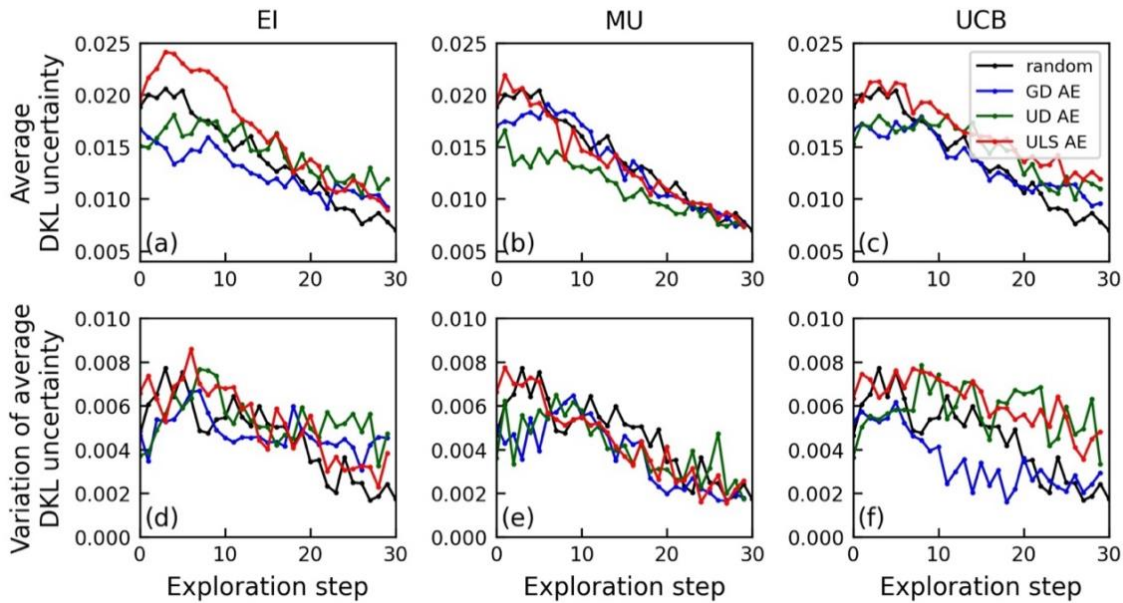

**FIG. 6.** Learning curves averaged across both the feature space and all experiments for (a-c) for DKL uncertainties, (d-f) variations of average DKL uncertainties over the experiment.

It's notable that the percentage of AE curves trapped in the "exploratory stagnation" scenario consistently remains within the 10-25% range across all selected seed point models (Table S1). This "exploratory stagnation" phase can extend for over 30 initial exploration steps and persists only when measurements are conducted at all potential locations in the vicinity of the trapping points. The choice of the sampling model used for seed point selection **does not** notably influence the trapping effect during the subsequent AE evolution.

## V. SEED POINTS INTERVENTIONS

The human interventions in the experimental loop can be employed to moderate the evolution of AE. In intervention, the exploration process driven by BO is halted, and instead, a few subsequent measurement locations are selected directly by the operator or by some alternative algorithm. Following this, the ongoing process is again reverted to BO's control.

We have studied the potential of the rVAE-based sampling models as the algorithms for microstructural choice in the interventions to prompt a transition from "exploratory stagnation" toward the "normal" scenario. Two different strategies – *regional exclusion* and *regional prioritizing* – have been developed for rVAE-based interventions.

The *regional exclusion* method involves an omission of specific regions within the rVAE latent space from consideration for the subsequent microstructural choice. To address "exploratory stagnation", this strategy involves excluding areas surrounding several previously chosen points in the latent space, associated with the trapping clusters. The number of these points as well as the size of the excluded regions around them are considered as the hyperparameters. The GD, UD, and ULS probabilistic models can be constructed based on the latent distribution for intervention sampling, ensuring a zero probability of point selection within the removed regions (Fig. 7b). These excluded areas correspond to early studied microstructures and those similar to them. So, the exclusion type of intervention enhances the exploration component in AE increasing the probability of finding another perspective microstructures for further investigation. From this perspective, the utilization of the UD model is most reasonable, because it prioritizes a diversity of chosen microstructure.

The *regional prioritizing* strategy focuses on prioritizing certain regions within the VAE latent space for the microstructural choice. This approach gives direct ability for the operator to influent the point selection scenario by choosing the promising regions for sampling in the latent or real spaces. When the operator selects a microstructure in real space, the algorithms identify and choose the area surrounding its latent representation. Afterward, the GD sampling models are constructed using the points inside the selected region and employing for the location choice during the interventions (Fig. 7c). The implementation of the VAE region selection instead of directly choosing the next measurement point enables a focus on a class of similar microstructures rather than a specific object.

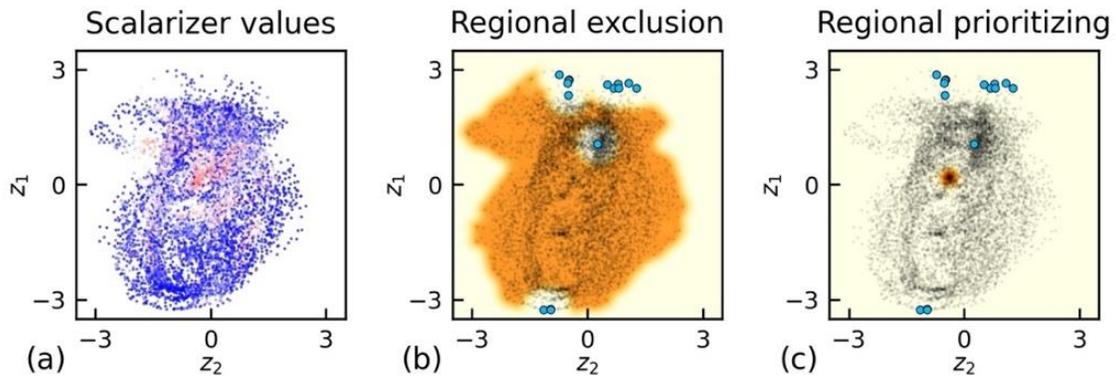

**FIG. 7.** rVAE latent distributions with (a) values of the scalarizer, probability densities for (b) regional exclusion, (c) regional prioritizing strategies. The blue points at (b, c) correspond to the already studied microstructures during the previous exploration steps.

The capability to prevent "exploratory stagnation" using seed point interventions was tested by the 15 experiment simulations for both regional prioritizing and regional exclusion approaches and EI, MU acquisition functions (Fig. S2). The each of DKL experiment simulations began from the 5 initial points, chosen with GD sampling model. After the first 20 exploration steps, each trajectory was bifurcated into two branches (Fig. 8a). In the first scenario, the prioritizing or exclusion intervention model guided the selection of 5 subsequent points before the continuation of AE exploration. In the second branch, AE exploration proceeded without any intervening actions. Within our simulations, 20% (3 out of 15) of experiments exhibit exploratory stagnation for branches without intervention for both considered acquisition functions (Table S2).

The **regional exclusion** intervention yields varying outcomes on the AE guided by the MU and EI. The interventions show ~67% efficiency for the MU acquisition function, where we note a transition back to the normal scenario immediately following the intervention for 2 of 3 stagnated experiments. Oppositely, in the case of the EI acquisition function, the number of learning curves exhibiting exploratory stagnation increased with the intervention. Upon more detailed analysis, it was found that among the 3 experiments showcasing exploratory stagnation without intervention, one was successfully alleviated from stagnation through regional exclusion. Interestingly, two additional experiments, which initially transitioned to the normal scenario without external influence, maintained stagnation upon intervention.

Figure 8(b-d) illustrates a successful instance of regional exclusion intervention for the MU acquisition function. At the initial stage of the experiment, we observe a smooth decline in average DKL uncertainty with the exploration step number associated with "exploratory stagnation" (Fig 8b). The seed points intervention leads to a sharp increase in both the variation (kernel sizes) and the average values of DKL uncertainty. The noticeable change in DKL uncertainty following the intervention aligns with the exploration typical of a "normal" evolution scenario in the experiment. The release effect is noticeable in the experimental trajectories. The point clusters identified prior to the intervention (Fig. 8c, blue points) have been supplanted by the points dispersed throughout the real and latent spaces following the intervention (Fig. 8c, red points). Important to note that in the branch without intervention, the experiment exhibits self-liberation after the 40 exploration steps, however, the intervention notably expedites this process.

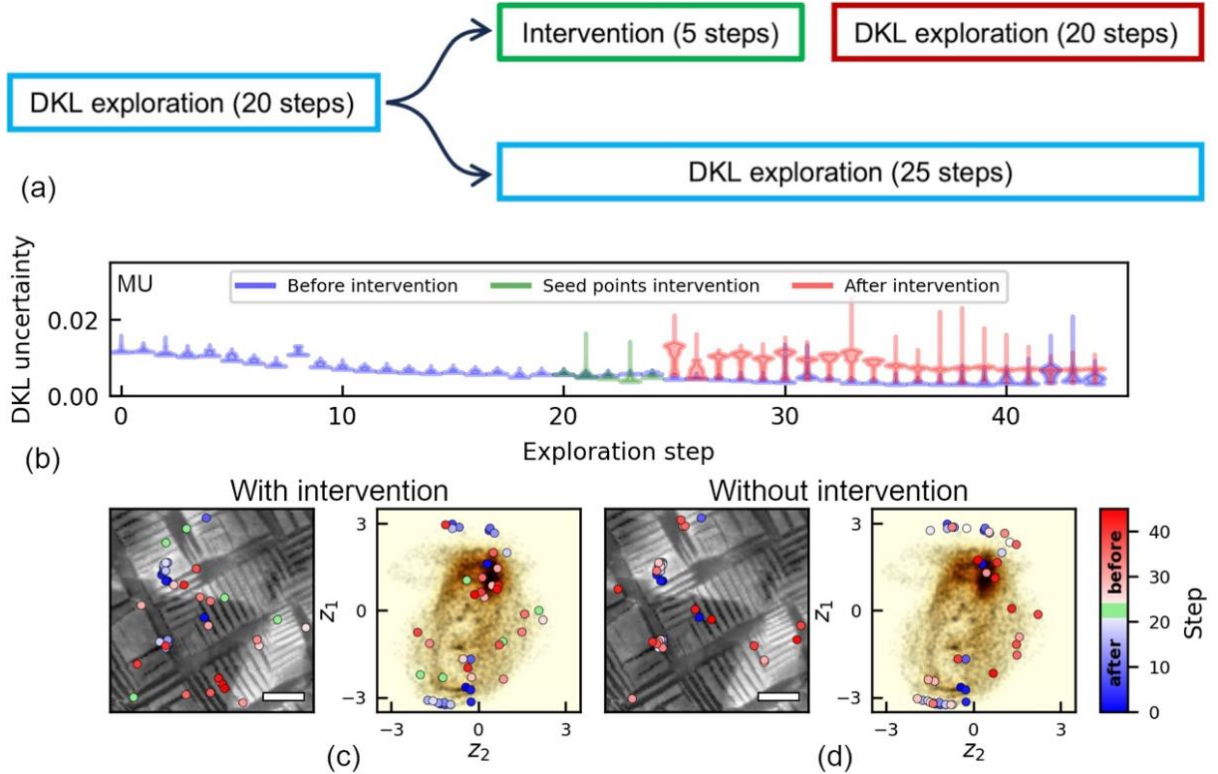

**FIG. 8.** Seed interventions in DKL AE: (a) experiments scheme; (b) learning curves for MU acquisition function with **regional exclusion** intervention and without any interruptions; experimental trajectories in real and rVAE latent spaces (c) with intervention and (d) without any intervening actions in DKL AE. The scale bar is 600 nm.

The efficiency of **regional prioritizing** was also estimated for the EI and MU acquisition functions. We examined two extreme cases in which the selected latent regions during the intervention exhibited scalarizer values much higher ("good" guess) or lower ("poor" guess) than those in the trapping regions. The interventions demonstrated 100% effectiveness in liberating the exploration process using the MU acquisition function for the "good" guess scenario. However, in the case of the "poor" guess scenario, the experiment was released in only one out of three cases. Similarly, with the EI acquisition function, the shift to the "normal" evolutionary scenario occurred exclusively when the scalarizer values in the prioritizing region were higher compared to those in the trapping region (Table S2). It highlights the primary difference between the predominantly exploitative and exploratory algorithms. For experiments guided by the MU acquisition function, the diversity of the selected microstructures directly impacts the "curiosity" level of the automated experiment. As a result, the introduction of unusual data during the intervention may promote experiment release even if the scalarizer values in the selected points are lower than in the points explored earlier. Oppositely, when employing the EI acquisition function, the value of the scalarizer at the selected locations outweighs the diversity of the chosen microstructures. Thus, trajectory release occurs only when an operator selects a more promising region for sampling compared to the regions where an experiment was trapped.

The instances of successful and unsuccessful interventions are demonstrated in Figure 9. In the scenario of a "good" guess, the introduction of intervention into AE exploration leads to an immediate rise in the learning curve's violin kernel sizes and average DKL uncertainty (Fig. 9a). Following the liberation, the released algorithm avoids clustering within the prioritizing region, opting instead to explore the entirety of the latent space (Fig. 9b). Conversely, interrupting the AE with a "poor" guess intervention does not affect the process. The learning curves for branches with and without intervention appear nearly identical (Fig. 9d). The points clustering remains consistently noticeable throughout the entire AE process (Fig. 9e).

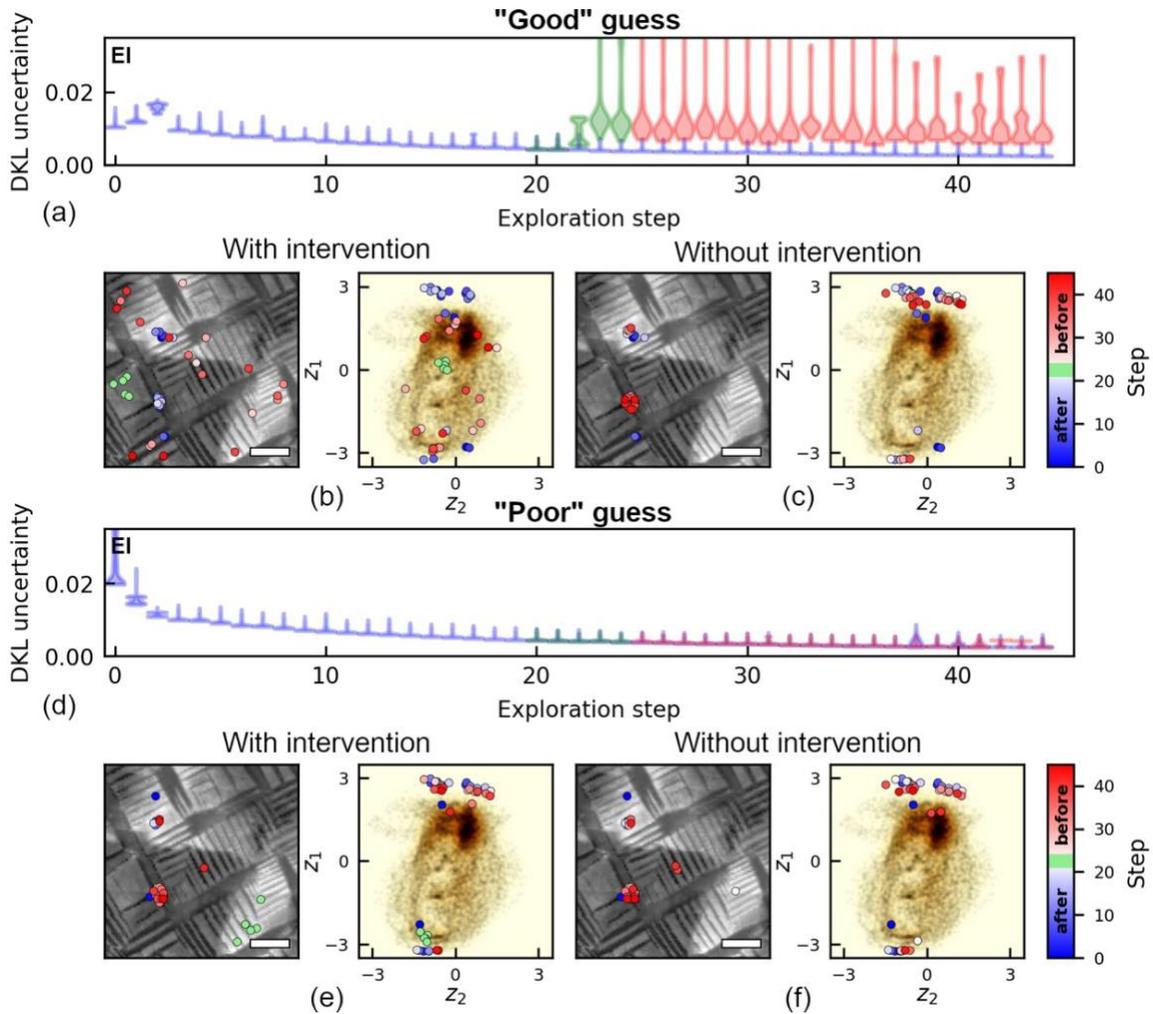

**FIG. 9.** Examples of DKL AE with **regional prioritizing** interventions: (a, d) learning curves represented by violin plots with branches prioritizing regions with higher and lower scalarizer values, respectively; experimental trajectories in real and rVAE latent spaces (b,e) with intervention and (c,f) without any intervening actions in DKL AE. The locations explored before the intervention are represented by blue points, after the intervention by red points, and during the intervention by green points. The scale bar is 600 nm.

## VI. CONCLUSION

To summarize, we have explored several strategies to employ the rVAE latent representation of the predefined structural data for the experiment initialization and seed point interventions in the AE process. We introduced three rVAE-based sampling models for the initialization. These models enable the selection of the most statistically valuable microstructures (GD model), prioritize microstructural diversity in sampling (UD model), or favor rare microstructures represented by points near the latent distribution edges (ULS model). The importance of the initialization and leveraging of the sampling policy depends on the diversity of the microstructures within the global image and selected acquisition function. An initial points selection plays an important role only at the beginning of the AE exploration.

At the same time, it was observed that the choice of a specific initialization model does not influence the probability of encountering "exploratory stagnation" in the AE evolution scenario. To counteract "exploratory stagnation," we introduced regional exclusion and regional prioritizing sampling models designed for intervention within the exploration loop. The regional exclusion approach is based on the UD model but with the elimination of trapping regions in the latent space from consideration. Consequently, the probability of selecting points from these regions is taken as zero. The regional prioritizing model allows for an operator to directly influent the AE process by choosing the specific region for sampling within rVAE latent space. The developed strategies were tested on the AE guided by EI (mainly exploitation) and MU (pure exploration) acquisition functions. We achieved a 70% effectiveness rate using the exclusion intervention strategy for the MU acquisition function. The release of the experiment trajectory with the EI acquisition function only happened when employing a prioritizing intervention strategy and the scalarizer values in the selected points are higher (in the case of maximization) compared to previously explored points. The regional prioritizing models with lower scalarizer values in the selected region yielded the least favorable outcomes – 0 out of 3 liberations for the EI function and 1 out of 3 for the MU function.

This work marks the next essential step in the design of robust workflows for automated experiments in microscopy. While the approaches and models detailed here were tested on SPM data, their principles can be readily adapted and applied to various other microscopy techniques.

## SUPPLEMENTARY MATERIAL

Supplementary material shows the impact of choices for initial measurements and adaptive interventions on the learning rates and efficiency of AE. Figure S1 illustrates the learning curves for experiments started with various initial sampling policies and guided by different acquisition functions. Table S1 shows the rate of 'exploratory stagnation' for various initial sampling policies and acquisition functions. Figure S2 displays the learning curves for experiments with various types of seed intervention. Table S2 provides information about the rate of 'exploratory stagnation' when various seed interventions are applied.

## ACKNOWLEDGMENTS

Y.L. acknowledges support by the Center for Nanophase Materials Sciences (CNMS), which is a US Department of Energy, Office of Science User Facility at Oak Ridge National Laboratory. S.V.K acknowledges support from the Center for Nanophase Materials Sciences (CNMS) user facility, project CNMS2023-B-02177, which is a US Department of Energy, Office of Science User Facility at Oak Ridge National Laboratory. The work was supported (S. V. K.) via start-up funding.

## AUTHOR DECLARATIONS

### Conflict of Interest

The authors have no conflicts to disclose.

### Author Contributions

**Boris N. Slautin**: Conceptualization (equal); Data curation (equal); Writing – original draft. **Yongtao Liu**: Conceptualization (equal); Data curation (equal); Writing – review & editing (equal). **Hiroshi Funakubo**: Resources. **Sergei V. Kalinin**: Conceptualization (lead); Supervision; Writing – review & editing (equal).

## DATA AVAILABILITY

The data that support the findings of this study are available from the corresponding authors upon reasonable request.

## REFERENCES

1. R. K. Vasudevan, K. Choudhary, A. Mehta, R. Smith, G. Kusne, F. Tavazza, L. Vlcek, M. Ziatdinov, S. V. Kalinin, and J. Hattrick-Simpers, MRS Commun. **9**(3), 821–838 (2019).

2. S. V. Kalinin, Y. Liu, A. Biswas, G. Duscher, U. Pratiush, K. Roccapriore, M. Ziatdinov, and R. Vasudevan, arXiv:2310.05018 (2023).

3. M. Abolhasani, and E. Kumacheva, Nature Synthesis **2**(6), 483–492 (2023).

4. E. Stach, B. DeCost, A. G. Kusne, J. Hattrick-Simpers, K. A. Brown, K. G. Reyes, J. Schrier, S. Billinge, T. Buonassisi, I. Foster, C. P. Gomes, J. M. Gregoire, A. Mehta, J. Montoya, E. Olivetti, C. Park, E. Rotenberg, S. K. Saikin, S. Smullin, V. Stanev, and B. Maruyama, Matter **4**, 2702–2726 (2021).

5. A. L. Ferguson, and K. A. Brown, Annu. Rev. Chem. Biomol. Eng. **13**, 25–44 (2022).

6. M. Ziatdinov, Y. Liu, K. Kelley, R. Vasudevan, and S. V. Kalinin, ACS Nano **16**(9), 13492–13512 (2022).

7. M. Seifrid, R. Pollice, A. Aguilar-Granda, Z. Morgan Chan, K. Hotta, C. T. Ser, J. Vestfrid, T. C. Wu, and A. Aspuru-Guzik, Acc. Chem. Res. **55**(17), 2454–2466 (2022).

8. Y. Li, L. Xia, Y. Fan, Q. Wang, and M. Hu, ChemPhysMater **1**(2), 77–85 (2022).

9. W. Gao, P. Raghavan, and C. W. Coley, Nat. Commun. **13**, 1075 (2022).

10. K. Choudhary, B. DeCost, C. Chen, A. Jain, F. Tavazza, R. Cohn, C. W. Park, A. Choudhary, A. Agrawal, S. J. L. Billinge, E. Holm, S. P. Ong, and C. Wolverton, NPJ Comput. Mater. **8**, 59 (2022).

11. M. A. R. Laskar, and U. Celano, APL Mach. Learn. **1**, 041501 (2023)

12. H. Tao, T. Wu, S. Kheiri, M. Aldeghi, A. Aspuru-Guzik, and E. Kumacheva, Adv. Funct. Mater. **31**(51), 2106725 (2021).

13. R. W. Epps, A. A. Volk, K. G. Reyes, and M. Abolhasani, Chem. Sci. **12**, 6025–6036 (2021).

14. M. Ahmadi, M. Ziatdinov, Y. Zhou, E. A. Lass, and S. V. Kalinin, Joule **5**, 2797–2822 (2021).

15. K. M. Roccapriore, O. Dyck, M. P. Oxley, M. Ziatdinov, and S. V. Kalinin, ACS Nano **16**(5), 7605–7614 (2022).

16. K. M. Roccapriore, S. V. Kalinin, and M. Ziatdinov, Adv. Sci. **9**, 2203422 (2022).

17. Y. Liu, K. P. Kelley, R. K. Vasudevan, W. Zhu, J. Hayden, J. Maria, H. Funakubo, M. A. Ziatdinov, S. Trolier-McKinstry, and S. V. Kalinin, Small **18**(48), 2204130 (2022).

18. Y. Liu, A. N. Morozovska, E. A. Eliseev, K. P. Kelley, R. Vasudevan, M. Ziatdinov, and S. V. Kalinin, Patterns **4**(3), 100704 (2023).

19. Y. Liu, K. P. Kelley, R. K. Vasudevan, H. Funakubo, M. A. Ziatdinov, and S. V. Kalinin, Nat. Mach. Intell.**4**, 341–350 (2022).

20. S. V. Kalinin, R. Vasudevan, Y. Liu, A. Ghosh, K. Roccapriore, and M. Ziatdinov, Mach. Learn.: Sci. Technol. **4**, 023001 (2023).

21. J. C. Thomas, A. Rossi, D. Smalley, L. Francaviglia, Z. Yu, T. Zhang, S. Kumari, J. A. Robinson, M. Terrones, M. Ishigami, E. Rotenberg, E. S. Barnard, A. Raja, E. Wong, D. F. Ogletree, M. M. Noack, and A. Weber-Bargioni, npj Comput Mater **8**, 99 (2022).

22. A. Krull, P. Hirsch, C. Rother, A. Schiffrin, and C. Krull, Commun. Phys. **3**, 54 (2020).

23. B. Alldritt, P. Hapala, N. Oinonen, F. Urtev, O. Krejci, F. F. Canova, J. Kannala, F. Schulz, P. Liljeroth, and A. S. Foster, Sci. Adv. **6**, eaay6913 (2020).

24. O. M. Gordon, J. E. A. Hodgkinson, S. M. Farley, E. L. Hunsicker, and P. J. Moriarty, Nano Lett. **20**, 7688–7693 (2020).

25. M. Rashidi, and R. A. Wolkow, ACS Nano **12**(6), 5185–5189 (2018).

26. J. Sotres, H. Boyd, and J. F. Gonzalez-Martinez, Nanoscale **13**, 9193–9203 (2021).

27. A. McDannald, M. Frontzek, A. T. Savici, M. Doucet, E. E. Rodriguez, K. Meuse, J. Opsahl-Ong, D. Samarov, I. Takeuchi, W. Ratcliff, and A. G. Kusne, Appl. Phys. Rev. **9**, 021408 (2022).

28. M. M. Noack, P. H. Zwart, D. M. Ushizima, M. Fukuto, K. G. Yager, K. C. Elbert, C. B. Murray, A. Stein, G. S. Doerk, E. H. R. Tsai, R. Li, G. Freychet, M. Zhernenkov, H.-Y. N. Holman, S. Lee, L. Chen, E. Rotenberg, T. Weber, Y. Le Goc, M. Boehm, P. Steffens, P. Mutti, and J. A. Sethian, Nat. Rev. Phys. **3**, 685–697 (2021).

29. S. Maruyama, K. Ouchi, T. Koganezawa, and Y. Matsumoto, ACS Comb. Sci. **22**(7), 348–355 (2020).

30. Y. Liu, K. Roccapriore, M. Checa, S. M. Valleti, J.-C. Yang, S. Jesse, and R. K. Vasudevan, arXiv:2312.10281 (2023).

31. R. K. Vasudevan, S. M. Valleti, M. Ziatdinov, G. Duscher, and S. Somnath, Adv. Theory Simul. **6**(11), 2300247 (2023).

32. B. Huang, Z. Li, and J. Li, Nanoscale **10**, 21320–21326 (2018).

33. N. Borodinov, S. Neumayer, S. V. Kalinin, O. S. Ovchinnikova, R. K. Vasudevan, and S. Jesse, npj Comput Mater **5**, 25 (2019).

34. V. Kocur, V. Hegrová, M. Patočka, J. Neuman, and A. Herout, Ultramicroscopy **246**, 113666 (2023).

35. A. Biswas, Y. Liu, N. Creange, Y.-C. Liu, S. Jesse, J.-C. Yang, S. V. Kalinin, M. A. Ziatdinov, and R. K. Vasudevan, npj Comput Mater **10**, 29 (2024).

36. A. G. Wilson, Z. Hu, R. Salakhutdinov, and E. P. Xing, Proceedings of the 19th International Conference on Artificial Intelligence and Statistics **51**, 370–378 (2016).

37. Y. Liu, J. Yang, R. K. Vasudevan, K. P. Kelley, M. Ziatdinov, S. V. Kalinin, and M. Ahmadi, J. Phys. Chem. Lett. **14**(13), 3352–3359 (2023).

38. Y. Liu, M. A. Ziatdinov, R. K. Vasudevan, and S. V. Kalinin, Patterns **4**(11), 100858 (2023).

39. S. Jesse, H. N. Lee, and S. V. Kalinin, Rev. Sci. Instrum. **77**(7), 073702 (2006).

40. M. Valleti, Y. Liu, and S. Kalinin, arXiv:2303.18236 (2023).

41. D. P. Kingma, and M. Welling, Found. Trends Mach. Learn. **12**(4), 307–392 (2019).

42. S. V. Kalinin, J. J. Steffens, Y. Liu, B. D. Huey, and M. Ziatdinov, Nanotechnology **33**(5), 055707 (2022).

43. S. V. Kalinin, O. Dyck, S. Jesse, and M. Ziatdinov, Sci. Adv. **7**(17), eabd5084 (2021)